\documentclass[conference]{IEEEtran}
\IEEEoverridecommandlockouts
\usepackage{cite}
\usepackage{amsmath,amssymb,amsfonts}
\usepackage{algorithmic}
\usepackage{graphicx}
\usepackage{pgfplots}
\usepackage{stfloats}
\usepackage{subfig}
\usepackage{textcomp}
\usepackage{xcolor}
\def\BibTeX{{\rm B\kern-.05em{\sc i\kern-.025em b}\kern-.08em
    T\kern-.1667em\lower.7ex\hbox{E}\kern-.125emX}}

\usepackage{hyperref}
\usepackage[T1]{fontenc} 
\usepackage[utf8]{inputenc}
\usepackage{verbatim}
\usepackage{float}
\usepackage{multirow}
\usepackage{booktabs}
\usepackage{tikz}
\newcommand*\circled[1]{{\scriptsize\tikz[baseline=(char.base)]{\node[shape=circle,draw,inner sep=2pt] (char) {#1};}}}

\usepgfplotslibrary{groupplots}
\usetikzlibrary{matrix,positioning}
\pgfplotsset{compat=newest}

\newcommand{\ARXIV}[1]{\textcolor{black}{#1}}
\newcommand{\CHANGE}[1]{\textcolor{black}{#1}}

\begin{document}

\title{Introducing Tales of Tribute AI Competition
\thanks{This work was supported by the National Science Centre, Poland under project number  2021/41/B/ST6/03691.}
}

\author{

        
    \IEEEauthorblockN{Jakub Kowalski, Rados{\l}aw Miernik, Katarzyna Polak,
    Dominik Budzki, Damian Kowalik}
    \IEEEauthorblockA{
        \textit{Institute of Computer Science, University of Wroc{\l}aw}\\
        Wroc{\l}aw, Poland \\
        Email: \{jakub.kowalski,radoslaw.miernik\}@cs.uni.wroc.pl; \{299135,314625,315984\}@uwr.edu.pl
    }
    
}

\maketitle

\begin{abstract}
This paper presents a new AI challenge, the Tales of Tribute AI Competition (TOTAIC), based on a two-player deck-building card game released with the High Isle chapter of The Elder Scrolls Online. Currently, there is no other AI competition covering Collectible Card Games (CCG) genre, and there has never been one that targets a deck-building game. Thus, apart from usual CCG-related obstacles to overcome, like randomness, hidden information, and large branching factor, the successful approach additionally requires long-term planning and versatility. The game can be tackled with multiple approaches, including classic adversarial search, single-player planning, and Neural Networks-based algorithms. This paper introduces the competition framework, describes the rules of the game, and presents the results of a tournament between sample AI agents. 
\end{abstract}


\section{Introduction}

Games were always used as a testbed for Artificial Intelligence and as a domain where many new approaches and algorithms were showcased \cite{Togelius2016AIResearchers}. 
Thus, game-based AI competitions are an excellent opportunity to stimulate research by tackling challenging problems within relevant and just fun environments and with a pinch of competitiveness.

Minimax was a breakthrough for Chess \cite{Campbell2002Deep}, Temporal-Difference Learning for Backgammon \cite{tesauro1994td}, Monte Carlo Tree Search for General Game Playing \cite{finnsson2008simulation}, Deep Reinforcement Learning for Atari 2600 games \cite{mnih2015human}; last two methods combined for Go \cite{silver2016mastering}.
These grand achievements were accompanied by a variety of other smaller developments and improvements in algorithms related to games of different types.
Each genre requires a specific approach and pushes the limits of known algorithms to meet the game style.

Collectible Card Games (CCG), e.g., Magic: The Gathering, Hearthstone, Pokémon Trading Card Game, Yu-Gi-Oh!, are characterized by their high complexity, large branching factor, randomness,  partial observability, and dynamic metagame \cite{hoover2020many}.
They usually contain a vast amount of cards, and although only a few are present in each match, the proper algorithm has to consider the appearance of every one of them.
Unlike classical board games, each turn is complex, and consists of many smaller actions that may influence what follow-ups would be possible.
Moreover, an action can have a nondeterministic outcome, so even the task of computing a set of legal moves for a single turn cannot be simply done at the turn's start.

The usual source of randomness in the form of shuffled piles of cards inherited from classical card games also applies.
Critical information is hidden from the player, including the opponent's hand, deck, and even own draws, so every action is taken based on speculations, and its probability of success has to be weighed.
Last but not least, metagame allows one to reason about the opponent's strategy and possible plays based on the knowledge of successful playstyle archetypes.

Usually, in games like Hearthstone, the deckbuilding phase is not a part of the match but takes place before. Each player prepares their deck using cards from their collection and play a game with an already prepared (yet unknown to the opponent) list. 
For example, in Hearthstone AI Competition \cite{Dockhorn2019Introducing}, there were two tracks: \textit{Premade Deck Playing}, where agents were forced to play decks prepared by the organizers, and \textit{User Created Deck Playing}, where the competitors had to pair their agent with an arbitrarily chosen deck.
Initial editions of Strategy Card Game AI Competition \cite{Kowalski2023SummarizingStrategy} featured another approach, \emph{arena mode}, where the agent was forced to build a deck before each match, within a repeated process of choosing a single card out of three options. 
Although this scenario poses some deckbuilding aspect on the agent, it is still a part that is separated from the main play and can be relatively easily scripted or even hardcoded.

In games like Dominion, which started an entirely new subgenre of deck-building card games, things work differently \cite{Heijden2014AnAnalysis}. 
Each player begins with a small, predefined deck of cards that they improve by purchasing cards from a common supply that is randomized and varies from game to game.
Thus, the player's deck is gradually built as the game progresses, with every turn decision about which cards to buy, and knowing that each owned card is usually used multiple times as the deck rotates.
This novelty shifts the players' attention from just using cards to proper deck management and usually puts a heavier accent on long-term planning that has to be paired with the flexibility to react to changes caused by randomness or the opponent properly.

For this reason, we introduce a new CCG-based AI contest: Tales of Tribute AI Competition. 
We believe it is a great addition to the set of currently available benchmarks, introducing the right amount of novelty compared to the past ones, and that it will become an opportunity to boost algorithm research. 

\section{Related AI Competitions}

Over recent years, two academic CCG-related competitions were held that are important reference points for our challenge, as we learned on their experience.
Neither is still organized.

\subsection{Hearthstone AI Competition}

Hearthstone is a popular digital CCG that has a large competitive scene.
The Hearthstone AI Competition \cite{Dockhorn2019Introducing} was held three times (2018--2020) at IEEE Conference on Games. It has been well-received by the AI and gaming communities, attracting a steady number of participants on two tracks.
As mentioned before, one track required using a deck prepared by the organizers (some of them, but not all, were known before), while the other allowed to bring own deck to the table.

Hearthstone AI Competition used a SabberStone -- a Hearthstone simulator written in C\# .Net Core that claimed to implement 98\% of the base cards from the game (as of 2019; development stopped then).
To create an agent, a participant had to implement an \texttt{AbstractAgent} class and a number of methods, including the main one that receives a game state and returns an action to perform. The time limit to finish the whole turn was 30 seconds.
Sabberstone provides means to simulate the game randomly and introduces dummy cards instead of the opponent's hand to handle partial observability.
While developing our competition framework, we tried to include the best of SabberStone, which served as our inspiration. 

Winning strategies of the submitted agents include MCTS, Rolling Horizon Evolution, Pruned BFS, and Dynamic Lookahead algorithm, usually paired with a state evaluation function.

\subsection{Strategy Card Game AI Competition}

Legends of Code and Magic (LOCM) \cite{Kowalski2023SummarizingStrategy} is a small programming game developed for the official CodinGame contest in 2018, where it attracted over 2,000 participants\footnote{\url{https://www.codingame.com/contests/legends-of-code-and-magic-marathon}} .
Later, the game was slightly extended and started as an academic competition (2019--2021), then extended again (2022).
Overall, six editions were organized, hosted by IEEE Conference on Games and IEEE Congress on Evolutionary Computation.

LOCM was designed as a toy-size problem, with the AI research in mind, and to encourage trying new ideas.
The game contains 160 cards, and all cards’ effects are deterministic, thus nondeterminism is introduced only via the ordering of cards and an unknown opponent’s deck. The game is played in the \textit{fair arena} mode, i.e., before every game, both players create their 30-card decks secretly from the symmetrical yet limited card choices.
The last edition introduced \textit{Constructed Mode}, where an agent can construct their deck by picking all cards at once from a given pool, but in each game, this pool is different as it contains randomly generated cards. 

The contest did not pose language restrictions, as long as the agent ran on a Unix system. 
The communication protocol based on the standard input-output.
It operated on significantly smaller timeframes: depending on the edition, it was either 100 or 200 milliseconds per standard turn.
Winning approaches were mostly based on various search algorithms (Flat MC, MCTS, Minimax) accompanied by handmade heuristics. However, the last edition was dominated by neural network approaches.

\section{Related Research}

It has been pointed out that Collectible Card Games pose many interesting challenges for AI research \cite{hoover2020many}.
The most straightforward (to define, not necessarily to achieve) goal is to successfully play the game.

A variety of approaches were proposed, though due to the stochastic nature of the game, many took the form of MCTS \cite{Browne2012ASurvey} enhancements.
Some work focuses on learning weights for evaluation functions that are further used in search algorithms.
An approach using competitive coevolutionary optimization for this purpose, and using it in a greedy one-step look-ahead algorithm (2nd place in 2018 Hearthstone AI Competition) is presented in \cite{garcia2020optimizing}.
In \cite{Montoliu2020efficientHeuristic}, the evaluation for Online Evolutionary Planning AI playing LOCM was trained using N-Tuple Bandit Evolutionary Algorithm.
LOCM-based study in \cite{Miernik2022EvolvingEvaluation} analyzes the influence of representation and the choice of opponent used to test the model on the quality of learned heuristics.
\ARXIV{Simulation-based approaches often require some means for reducing search and action space.}
Combination of many interesting enhancements, including state abstraction, optimized Information Set MCTS, and sparse sampling was described in \cite{Choe2019EnhancingMonteCarlo} (Hearthstone AI Competition winner 2019).
\ARXIV{Another Hearthstone-based study, \cite{zhang2017improving}, extends Determinized MCTS with bucketing chance nodes and high-level playouts, also testing neural networks for learning card playing policy.}
Neural networks have proven to work really well in the CGG domain, e.g., used as an evaluation function for MCTS search \cite{swiechowski2018improving}. 
Recent most spectacular successes include double-winning the last edition of Strategy Card Game AI Competition \cite{Xi2023MasteringLOCM} and winning against the top 10 human player of the official Hearthstone League in China \cite{Xiao2023MasteringHS}.
\CHANGE{AI approaches for Dominion features similar methods, including MCTS \cite{jansen2014ai} and Deep Reinforcement Learning \cite{Gerigk_Engels_2023}.}

An interesting topic regarding CCGs is deck building, the task of preparing a combination of cards that, when paired with an AI agent, will be able to win against a variety of opponents consistently.
The usual approach for this task is to use some form of evolutionary algorithms (EA) \cite{garcia2016evolutionary}.
An approach tailored to the arena game mode in LOCM, extending EA with active genes to improve learning efficiency, was described in \cite{Kowalski2020EvolutionaryApproach}.
In \cite{Fontaine2019Mapping}, a modification of the MAP-Elites algorithm that introduced sliding grid cell boundaries was shown to discover high-performing Hearthstone decks.

\ARXIV{A related topic, touching the metagame, is game balancing.
Our aim is to ensure that the entire set of available cards does not degenerate to a single winning deck.
Some study, focusing on finding overperforming cards, can be found in \cite{Silva2019evolving}.
In \cite{mahlmann2012evolving}, a genetic algorithm is used to provide Dominion decks that result in balanced matches regardless of player skill differences.}

\section{Tales of Tribute Card Game}

Tales of Tribute (ToT) is a two-player deck-building card game released in 2022 as an expansion to the popular MMO RPG game \textit{The Elder Scrolls Online} (ESO), developed by ZeniMax Online Studios and published by Bethesda Softworks.

\begin{figure}[t]
    \centering
    \includegraphics[width= \columnwidth]{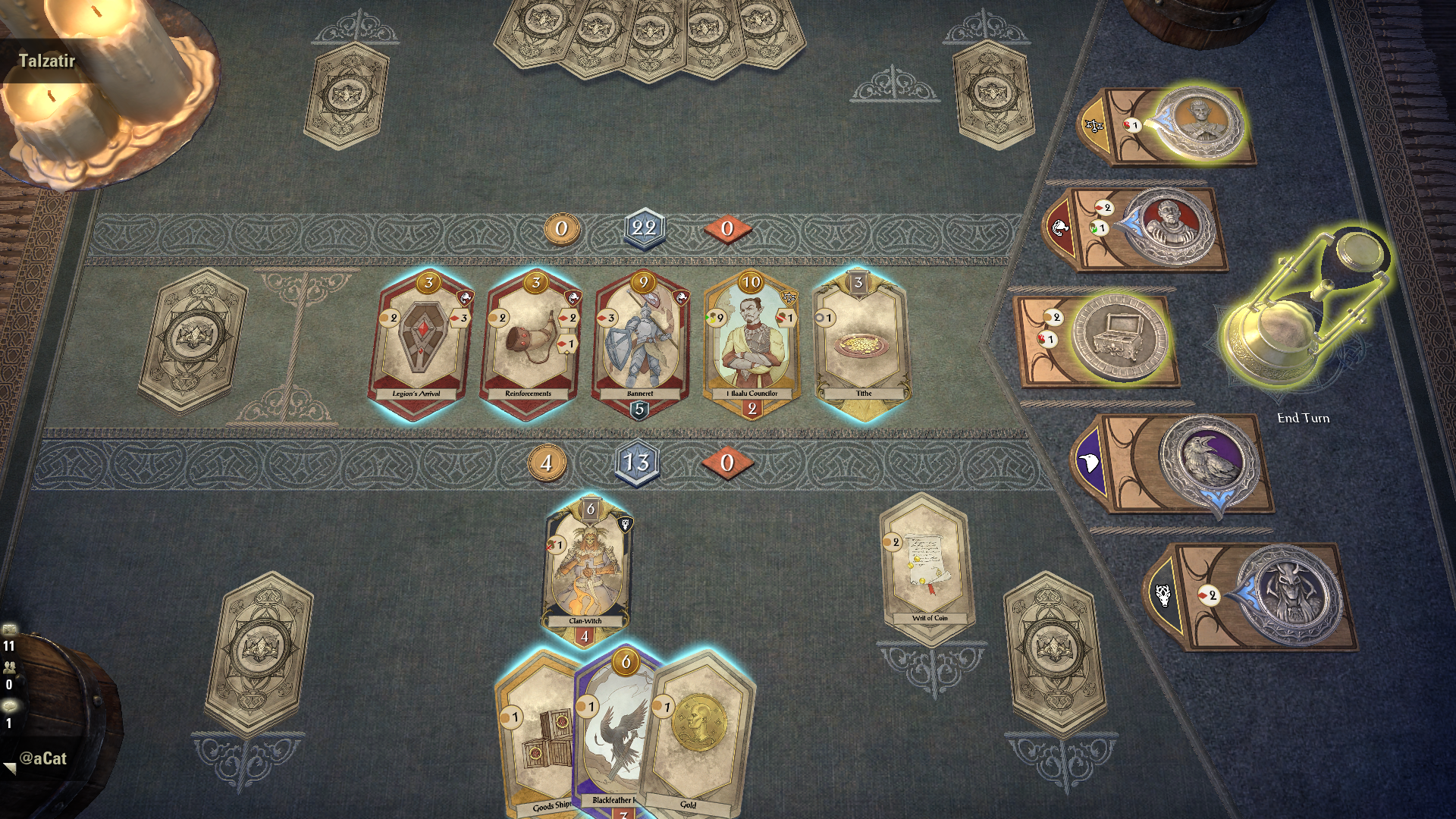}
    \caption{\ARXIV{Tales of Tribute in The Elder Scrolls Online.}}
    \label{fig:esoscreen}
\end{figure}

It features several decks of cards that differ in playing style. Each deck is represented by a \emph{patron}.
These decks can be acquired during ESO gameplay, and playing ToT is related to various ESO quests and achievements.
\ARXIV{Figure~\ref{fig:esoscreen} shows the screen of the original game.}

\subsection{Game Rules}

\begin{figure*}[t]
    \centering
    \includegraphics[width=0.95\textwidth]{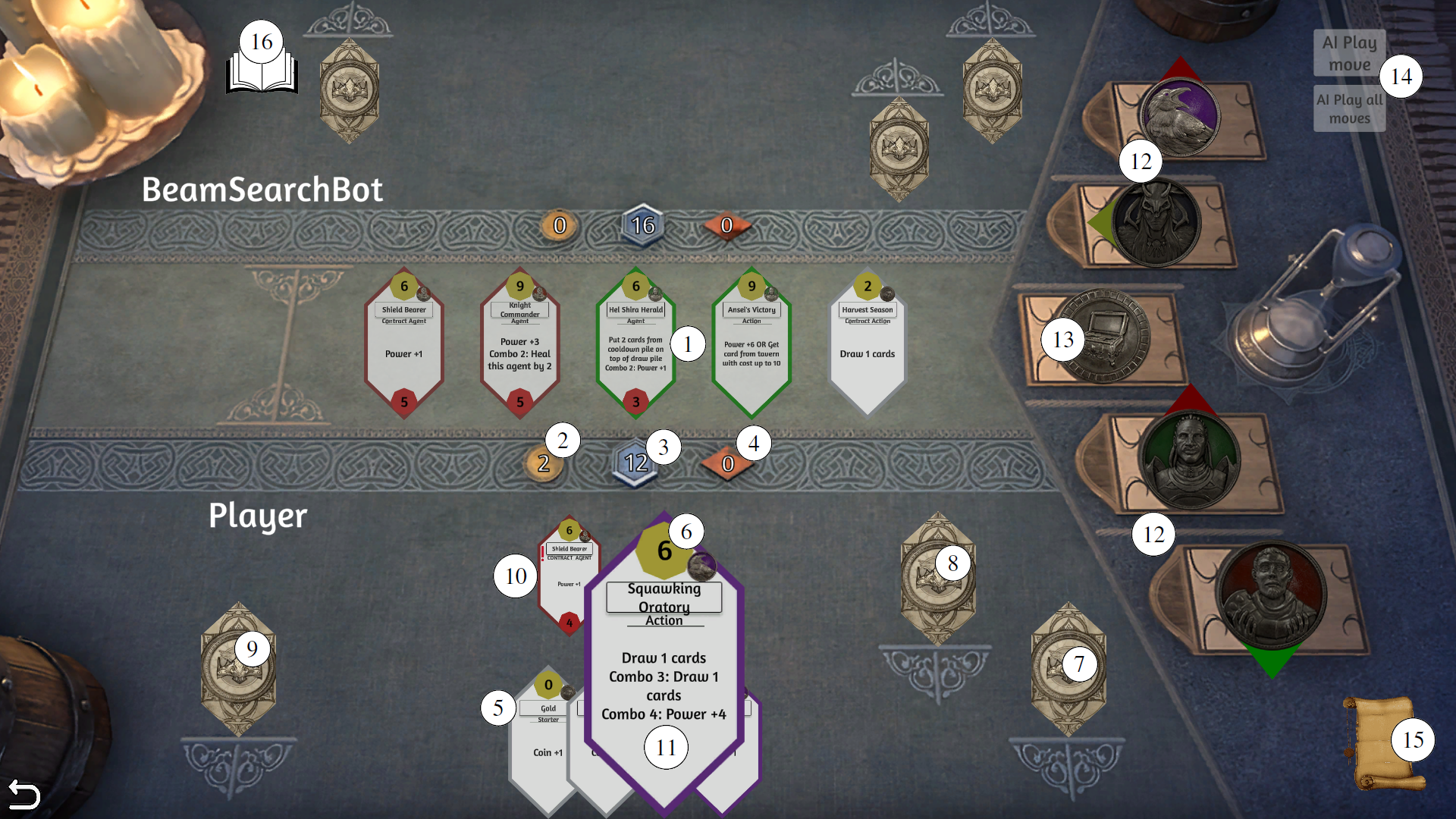}
    \caption{Graphical user interface of the game within Scripts of Tribute framework.}
    \label{fig:sotscreen}
\end{figure*}

Here, we present a compact description of the nearly-complete Tales of Tribute rules. 
Figure~\ref{fig:sotscreen} shows how the game looks within Scripts of Tribute GUI.
All interface elements are described below and referenced using numbers (e.g., \circled{1}).

The competition-ready Scripts of Tribute release is compatible with Tales of Tribute from ESO PC/Mac Patch 9.2.10 (25.02.2024) and contains seven out of eleven patrons available in ESO (three of which were added after the release of the game). 
All cards are fully upgraded. 
When enumerating a list of available patrons and keywords, only a subset used during the current (2024) edition competition is presented.

\subsubsection{Starting a match}

There are seven \emph{patron decks} available.
These are equivalent of suit of cards, colors, or classes from the other CCGs. Before the main part of the game, players choose 4 decks that will be played. The starting player chooses first and fourth, while the second player chooses second and third deck. There is also one neutral, so-called \emph{Treasury} deck that is present in every match.

\subsubsection{General playing rules}

Each player starts with a deck containing 10 cards: 4 are the \emph{starter} cards from the chosen decks, and six are the basic \textit{Treasury} cards called \emph{Gold}. 
The remaining deck cards (chosen ones and \textit{Treasury}) are shuffled and put to  \emph{tavern pile} face down. Top five cards are put in the middle and are visible to both players -- this is called the \emph{tavern} \circled{1}. Random 5 of players' cards are in their hands, and the rest is shuffled and put in their \emph{draw piles}.

There are three main resources in Tales of Tribute: \emph{coins}, \emph{prestige}, and \emph{power}. 
Coins \circled{2} are primarly used to buy cards from the \emph{tavern}. At the end of every turn, unused coins are lost. 
The main purpose of the prestige \circled{3} is to gather enough to win the game.
Power \circled{4} can be spent to attack the opponent’s \emph{agent cards} and unspent power automatically transitions to prestige at the end of the turn.

During their turn, a player can use cards in hand \circled{5} (any number of them, in any order). 
Using \textit{Gold} card from the initial hand gives +1 coin; starter cards usually give +1 coin or +1 power. 
The second player is handicapped, thus they start their first turn with 1 additional coin.
Each card has an associated coin cost \circled{6}, so by spending coins, one can buy cards from the \textit{tavern} (from the five available).
After purchasing a card it goes to \emph{cooldown pile} \circled{7}, and the top card from the \textit{tavern deck} goes to \textit{tavern} and can be bought.

Used cards go to the \emph{played pile} \circled{8} which, after the end of the turn, is merged with the \textit{cooldown pile}.
Next hand is drawn from the \textit{draw pile} \circled{9}. When it becomes empty, \textit{cooldown pile} is shuffled and becomes a new \textit{draw pile}.

Players alternate their turns, buying cards from the same \textit{tavern} and competing over the favor of same \textit{patrons}.

\subsubsection{Cards}

Usually, we have to pay the card's cost when buying a card from the \textit{tavern}. Using a card from our hand is always free. Every card belongs to exactly one deck (\textit{patron deck} or \textit{Treasury}).
There are several types of cards:
\begin{itemize}
\item \emph{starter}: Each patron 
has one type of a starter card. They are put to players' decks at the beginning of each match.

\item \emph{action}: Most common type of cards. When played from the hand, they trigger their effects and are put on \emph{played pile}, which is later merged with the \textit{cooldown pile}.

\item \emph{contract action}: They are played automatically as soon as they are acquired, trigger their effects and go to \textit{played pile}. However, at the end of the turn, they are not shuffled into \textit{cooldown pile}, instead they are removed from the match.

\item \emph{agent}: When played (from the hand), they are placed on the board, and remain there through the turns \circled{11}. In the following turns, a player can use an active agent triggering its effects, just as they would play another action.
Every player can have at most 7 active agents. Summoning a new agent trigger its effects but the card does not stay on board.
Agent cards have an additional \emph{health} statistic. During their turn, players can attack opponent's agents spending power, reducing agent's health by $\mathrm{min}(\mathrm{power}, \mathrm{agentHealth})$. When agent's health reaches 0, it is moved to the \textit{cooldown pile}.
Attacking opponent's agent is voluntary. However, some agents have \emph{taunt}, which automatically uses opponent's power to attack them, preventing its conversion to prestige.

\item \emph{contract agent}: They are played automatically as soon as they are acquired from the tavern. Otherwise, they behave like a standard agent with the exception that when destroyed they are removed from the match entirely.
\end{itemize}

\subsubsection{Combo Effects}

Using a card (action from hand, or active agent on the board) triggers its \emph{play effect}. However, most of the cards have additional effects that are set under the \emph{combo $n$} condition \circled{11}. These effects are triggered when $n$ cards from the same deck were used during the same turn. The card itself contributes to the counter, so \emph{play effect} is semantically equivalent to \emph{combo 1}.

When calculating a combo counter for cards from a given deck, the order of playing its cards and playing other deck's cards in between does not matter.
If a player plays a card from a deck, and there are already 2 cards in their played pile from the same deck, all the \emph{combo 3} effects will trigger: the one on the currently played card (if exists), and these on the cards in the \textit{played pile} (if exist).

\subsubsection{Keywords}

The possible effects of playing cards are described by a set of \emph{keywords}. The semantics  of each keyword is generally simple, and for each play/combo effect at most two keywords are used. Keywords may be merged by either \textbf{and} operator (effects of both keywords are applied) or \textbf{or} operator (player decides which effect applies).
Available keywords are summarized in Table~\ref{tab:keywords}.

\begin{table}
\begin{center}
    \caption{Keyword effects.}
    \label{tab:keywords}
    \begin{tabular}{ll}
    \toprule
 Keyword & Triggering effect \\ 
 \midrule\midrule
 \verb|ACQUIRE| $n$ & Acquire 1 card from the \textit{tavern} with a cost up to $n$ \\ \midrule
 \verb|COIN| $n$ & Gain $n$ coins \\ \midrule
  \multirow{3}{*}{\texttt{CREATE} $n$} & \ARXIV{Create $n$ \textit{Summerset Sacking} cards} \\ 
  & \ARXIV{(+1 Prestige, Combo 2: +1 Prestige, Combo 3: +1 Coin)} \\ 
  & \ARXIV{and place in cooldown pile} \\ \midrule 
 
 \verb|DESTROY| $n$ & Destroy up to $n$ of your cards that are in play \\ \midrule
 \multirow{2}{*}{\texttt{DISCARD} $n$} & Opponent discards $n$ cards from their hand \\
 & at the start of their turn \\ \midrule
 \verb|DRAW| $n$ & Draw $n$ cards from the \textit{draw pile}\\ \midrule
 \verb|HEAL| $n$ & Heal this agent for $n$ health \\ \midrule
 \verb|KNOCKOUT| $n$ & Set health of $n$ opponent's active agents to 0 \\ \midrule
 \verb|OPPLOSEPR| $n$ & Opponent loses $n$ prestige \\ \midrule
 \verb|PATRON| $n$ & Additional $n$ \textit{patron} activations this turn \\ \midrule
 \verb|POWER| $n$ & Gain $n$ power \\ \midrule
 \verb|REPLACE| $n$ & Replace up to $n$ cards from the \textit{tavern} \\ \midrule
 \verb|RETURN| $n$ & Return $n$ cards from cooldown to the top of your draw \\ 
    \bottomrule
    \end{tabular}
\end{center}
\end{table}

\subsubsection{Patrons}
Each deck in Tales of Tribute is represented by a \textit{patron} and the four patrons belonging to chosen decks are present during the game \circled{12}.
Patrons have a status visualized by an arrow pointing from the patron.
They can be neutral or favor one of the players.
All patrons start neutral.

Every patron has a special effect that benefits a player, but its activation comes with a cost.
Usually, activating a patron shifts their status one step towards a player. 
By default, a player can make one patron call per turn, but some card effects increase this counter.
\emph{Treasury} \circled{13} is a special patron, present in all matches. It does not have a status, yet its activation counts normally, decreasing counter by one.
Available patrons and their effects are summarized in Table~\ref{tab:patrons}.

\subsubsection{Winning}

There are a few ways to win. 
The first one is to earn the favor of all four patrons.
If a player achieves that in any moment of their turn, they immediately win, disregarding all other circumstances.

Other than that, when one player gains at least 40 prestige, the opponent needs to beat their score next turn to stay in the game. The match goes into a ``sudden death'' mode. The player who fails to match their opponent’s prestige in their turn loses. 
The upper limit is 80 prestige. The player who reaches this limit first wins after the turn ends.

For safety reasons, we included a hard limit of 500 turns, not present in the original game, resulting in a draw.
In practice, it is reached only by extremely naive agents (e.g., random).

\subsection{Playing strategy}

The starting goal is to build the economy, mainly by using Treasury patron to change \textit{Gold} cards to \textit{Writ of Coins} (giving +2 coins instead of +1). 
Buying average cards from the \textit{tavern} is a risk of revealing good cards for the opponent to buy, so usually only top cards are bought during this stage.

With a stable income allowing to buy 7+ cost cards every turn, the main deckbuilding phase begins. Buying good, possibly expensive, cards with a focus on one or two decks increases the probability of triggering high-level combo effects. These probabilities can be turned into certainties when combined with deck thinning (destroying owned weak cards) and additional card draw.

At some point, winning seem to be within reach. It is time to abandon buying cards that will not be of immediate use, and even sell cards for power if possible.
Focus should be shifted from coins to power/prestige generation, to reach the 40 prestige threshold as fast as possible, with the help of specific patrons like \textit{Crows} or \textit{Hlaalu} to make a final leap.

Please note that every match has its own characteristics, depending mainly on the patrons in play but also on what appears in the tavern. For example, usually matches with \textit{Hlaalu} have more easily accessible coins, while \textit{Rajin} games can be slower due to the \textit{Bewilderment} cards and decks' ability to decrease opponent prestige.

\section{Competition Framework}
We present Scripts of Tribute (SoT), a Tales of Tribute Simulator written in .Net C\#. To simplify the setup process, a Dockerfile has been added to the game engine repository. The environment created with this Dockerfile is consistent with the one used during the competition.
In this Section, we provide a brief overview of the framework and how to implement agents in it. For more information, please refer to the competition webpage\footnote{\url{https://github.com/ScriptsOfTribute}} and the thesis \cite{Budzki2023ImplementingTalesOfTribute}.

\renewcommand{\arraystretch}{0.92}
\begin{table}
\begin{center}
    \caption{Patron powers.}
    \label{tab:patrons}
    \begin{tabular}{ccl}
    \toprule
 Patron & Activation cost & Effect \\ 
 \midrule\midrule

\multirow{2}{*}{Ansei} & 2 Power & Activating player becomes favored;\\
 & (cannot favor player) & Favored player gain 1 coin at turn's start \\ \midrule
 \multirow{2}{*}{Crows} & All coins,  min.\ 1 & \multirow{2}{*}{Gain power equal to $\mathrm{coins} -1$}  \\
 & (cannot favor player) & \\ \midrule
 \multirow{2}{*}{Hlaalu} & \multirow{2}{*}{Sacrifice a card} & Gain prestige equal to  \\
 & & card's $\mathrm{cost} - 1$  \\ \midrule
  \multirow{2}{*}{Orgnum} & \multirow{2}{*}{1 / 2 / 3 Coins} & Gain power: 2 / 1 for every 6 / every 4 cards  \\
& & If favored create 1 Summerset Sacking card. \\ 
 & &  Improve Orgnum favor level. \\ \midrule
 \multirow{2}{*}{Pelin} & \multirow{2}{*}{2 Power} & Return an agent from cooldown  \\
 & & to the top of your deck \\ \midrule
 \multirow{2}{*}{Rajhin} & \multirow{2}{*}{3 Coins} & Place \emph{Bewilderment} (no effect) card   \\ 
 & & in opponent's cooldown pile\\  \midrule
 Red Eagle & 2 Power & Draw a card \\\midrule \midrule
 \multirow{2}{*}{Treasury} & \multirow{2}{*}{2 Coins} & Sacrifice 1 card and create\\
  & & 1 \textit{Writ of Coin} (gives +2 coins) \\ 
    \bottomrule
    \end{tabular}
\end{center}
\end{table}

\subsection{Implementing an Agent}

To implement an agent, a contestant has to implement a C\# class inheriting from the abstract \verb|AI| class and compile the code into a library. The resulting \verb|dll| file has to be placed in the right folder (OS-dependent) in order for GUI or Console Runner to be able to load this agent.

Implementing \verb|AI| class  requires overwriting three methods.
\verb|SelectPatron| is called two times when selecting patrons before the match. It receives a list of patrons still available to be picked and a number specifying pick turn. The method should return the identifier of the chosen patron.
\verb|Play| method is called each time the agent needs to make an action. It receives a \verb|GameState|, a list of legal actions, and should return one of them.
\verb|GameEnd| is called after a match and allows an agent to analyze the data from this match, stored in the given \verb|EndGameState| object.
There is a special \verb|Log| method the agent can call. It takes a string and appends it to the logs that can be later displayed by GUI or redirected by a Game Runner to a file.

\subsection{External Language Adapter}
Our game engine introduces a feature facilitating the development of bots in diverse programming languages. The communication between the game engine and bots is seamlessly achieved through pipes, ensuring a reliable data exchange mechanism. Bots receive input in the form of JSON payloads through the standard input stream, terminated by the \verb|EOT| marker. Upon parsing this JSON, bots communicate their moves back to the engine via the standard output stream, adhering to a predefined format documented separately.

This language-agnostic approach enhances the accessibility of the engine, promoting a collaborative and versatile environment for bot development. The documentation on the competition webpage comprehensively outlines the JSON structure for input, the required output format, and any game-specific rules, providing a solid foundation for participants in the tournament to leverage this functionality.

\subsection{Engine}

The \verb|GameState| object given in \verb|Play| method is the game state as perceived by the human player. It has access to e.g.\ agent's own \textit{draw pile}, but it is sorted lexicographically, not in the order of draw. Also, the opponent's hand and \textit{draw pile} are merged (we know their contents but not the order and which of the cards are currently in hand).
But our engine still allows to simulate a course of the game by using an \verb|ApplyState| method of \verb|GameState|. However, to deal with partial observability and randomness, the method requires the \verb|seed| parameter and switches the type of our state to \verb|SeededGameState|. This is a complete game state that has access to all piles with proper ordering but with the assumption that they were randomized using the given seed.

The computational model behind ToT is specific due to the fact that a single action, like playing a card, may cause a chain of effects (triggered combos), and each of these effects may require another action interrupting the chain and putting other choices to be resolved at the beginning (e.g. acquiring a card from the \textit{tavern} which requires choosing one of them).

For this reason, we make a strong distinction between normal actions and choices required by previous actions.
Possible action types are: \verb|ACTIVATE_AGENT|, \verb|ACTIVATE_PATRON|, \verb|ATTACK_AGENT|, \verb|BUY_CARD|, \verb|END_TURN|, \verb|MAKE_CHOICE|, and \verb|PLAY_CARD|.
Thus, the \verb|GameState| object knows if the current move to make is a pending choice and contains a queue of effects that have to be resolved after.

The engine is designed to be extendable. It allows adding new cards, new keywords, and even new patrons relatively easily. All cards are kept in a single \verb|cards.json| file, and a Python script generates C\# enums with card IDs.
Adding a new effect requires creating a new value in \verb|EffectType| enum and ensuring its proper behavior in several parts of the code.
Implementing a new patron requires making a class extending the abstract \verb|Patron| class.

\texttt{Game Runner} is a console application provided with the engine. It allows to load agents from \verb|dll| files (they have to be in the same directory as the runner) and run games between them.
It allows to specify, in particular, the number of matches to be played, seeds for these matches, and options for logging into files.
This tool will be used to run the competition.

\ARXIV{To present an estimation of the engine efficiency, we ran 1000 matches between two \textit{Random} agents using the Game Runner (on a single thread).
It took about 8.5 seconds on Intel Core i5-9300H, and 2.5 seconds on Intel Core i7-12700H, using C\# 7.0 and .Net Standard 2.1.
The memory overhead of the game engine is negligible.}

\subsection{GUI}
The graphical user interface of SoT is written using the Unity framework. 
It can be built from the available sources in the repository, but we publish most current versions as releases that contain built executables for Windows, Linux, and Mac OS. Figure~\ref{fig:sotscreen} shows what the game screen looks like.
The general layout of the original game is preserved, and most interaction behaviors were kept the same as in ESO ToT.

GUI allows playing a human versus AI agent match.
Before starting a match, there is a basic setup where we can choose an AI opponent, decide whether we go first or second, set the time limit per turn (for AI), and provide seed value if we aim to duplicate the order of \textit{tavern} cards and RNG calls from some previous match.
Then, there is a patron-picking phase, and afterward, the main part of the match begins. 
The user is always on the bottom of the screen.

The main purpose of GUI is to help with the development and debugging of agents. 
Thus, during the game, we play open cards and can see AI's hand, draw, etc.
To control when the agent is playing, we have two buttons available \circled{14}.
One calls an agent to make a single move and shows its effects on the screen; the other performs the entire turn at once.

There are two additional screens, absent in the original game.
First contains the detailed action history \circled{15}, listing all past actions: played cards, triggered effects, and other interactions.
This helps with understanding the game mechanics and checking the detailed consequences of one's actions.
The other screen shows the agent's logs in a copyable form \circled{16}.
If a \verb|Log| method is called in its class, it will be printed after each action taken by the AI.

\subsection{Submission and Evaluation}\label{sec:eval}

The agent submission requires sending its source code to the organizer's email, accompanied by a short description of the agent and optionally build and run instructions.

The evaluation is held via a round robin schedule between all players and selected sample agents (see Section~\ref{sec:agents}).
The games are played until the winrate ranking clearly stabilizes.
All matches are played in pairs, with switched starting positions for the same seeds (Tales of Tribute has a clear preference for the starting player). 

Agents are mainly expected to be written in C\#, but using other languages are supported via adapter is allowed. The time constraint for competition matches is 10 seconds per turn, and the memory used should not exceed 256 MB.
All details are described at the competition website.

\section{Experiments}\label{sec:agents}

To test our framework and validate its usefulness for competition purposes, we have developed several agents varying in strength 
and organized a tournament between them.

\subsection{Sample agents}

We briefly introduce some of the agents available in the competition repository. 
The agents presented here are described in more detail in the thesis\cite{Budzki2023ImplementingTalesOfTribute}.

\subsubsection{Random}
Plays actions picked uniformly at random out of the set of legal ones minus \verb|END_TURN|.
The agent ends their turn only when there are no other actions available.
In most cases, playing all the cards from a hand is a safe course of action. 
The play style of this agent seems to correspond to how NPCs play on the \textit{Novice} level in ESO.

A fully random agent, which can also uniformly finish the turn, is skipped in our comparison, as in the global tournament, it achieved a win ratio of less than 0.25\%.

\subsubsection{Max Prestige}
Maximizing prestige (and power, since it is usually converted to prestige) is the first straightforward strategy that comes to mind.
This agent simulates moves checking all paths of length up to two, and chooses the action with the highest sum of prestige and power.
It prefers  winning action if one is found.
The approach can be seen as a variant of a classic One-step Look Ahead (OSLA) algorithm, which proved to work well in various multi-action games.

\subsubsection{Patron Favors}
Most matches end with 40+ prestige, and it is easy to forget about the patron-based alternative win condition, which is usually tricky to achieve during a normal game.
Thus, this agent was created to test a simple strategy focusing on winning by favoring patrons. 

The agent prioritizes actions that lead to activating a patron that does not favor it. 
If it used all activations in this turn, it performs shallow search checking possibilities for another activation.
Otherwise, the agent plays as \emph{Random}.

\subsubsection{Max Agent}
As the cards of type agent are generally strong due to their more permanent presence on the board, we created a bot specialized in using them.
This bot starts its turn by randomly using all the \verb|ACTIVATE_AGENT| or \verb|PLAY_CARD| actions and then checks whether it can buy any agents, prioritizing regular agents over the contract ones.
Agents are sorted by their \emph{tier}, and the best one is selected for purchase.
If buying an agent is not possible, a random action is played (except \verb|END_TURN|).

This is the first of the algorithms that makes use of a \emph{card tier list} -- a ranking system that helps categorize cards based on their power, versatility, and synergies between them. It is a very popular approach that helps human players during the decision-making process, used both by beginners and experienced gamers.
Such tier lists can be made by using algorithms, as some researchers did for
Hearthstone. 
For the purpose of this experiment, we created a static list ourselves, subjectively dividing all cards into five tiers.

\subsubsection{Decision Tree}

The agent deterministically calculates the best action based on the actual state of the board.
It plays cards, beginning with those from \emph{Treasury}. Usually, it will play all the cards and then decide to make other moves.
The agent prioritizes buying cards that make their deck more powerful: high-tier cards or cards from decks it contains many cards from. 
Contract cards from the \textit{Treasury} deck are handled separately, based on the current state of the board, as their use is highly situational.
The agent manages patrons to prevent the opponent from winning and tries to win with them if it seems within reach.
It also has rules to ensure the patrons are used efficiently.
Choices are handled depending on their meaning (e.g., selecting a minimal amount of bad cards or a maximal amount of good cards), and the option that seems to be situationally better is chosen.

\subsubsection{Flat Monte Carlo}

The agent uses random simulations from the current root with a limited horizon -- it does not go beyond a random event, thus sticking to only what is surely known at this moment.
A random event is defined as drawing a card, an action that requires some choice, or buying a card after a new card appears in the \textit{tavern}. 
Each time the agent remembers the playout with the best heuristic value and uses it during subsequent subturns, until a random event occurs, when it starts new calculations. 
To reduce the number of early playout terminations, the probability of selecting \verb|END_TURN| when other moves are possible, was set to 0.1\%.

The reasoning style for this agent somewhat corresponds to how many people play this game -- as they try to figure out the best playouts with all information available right now.

All simulation-based bots use the same heuristic evaluation function, based on the set of features with assigned weights (both handcrafted and tuned using an evolutionary algorithm).
The function considers the amount of power and prestige, as well as the level of patrons' favoritism. If the agent's prestige is smaller than 30 (subjective threshold between the middle and endgame phases) it also takes into consideration: tiers of agent-type cards on the board (own and opponent's), tiers of cards in own deck, and amount of cards from the same deck, and penalties for cards left in the tavern, that are of high tier or that may suit the opponent deck composition.

\subsubsection{MCTS}
This agent uses a slightly modified version of the classic MCTS \cite{Browne2012ASurvey}.
It simulates only its own turn, and based on the state after the playout, assigns a heuristic score to this simulation.
It uses guided playouts, greedily choosing the action that leads to the state with the highest value.

The agent also uses modified UCT, taking maximum instead of average. Although it seems counterintuitive, as there are random events during our turn, and this way we do not properly estimate probabilities of such events, this behavior proved to work better during the tests.

\subsubsection{Beam Search}

Classic best-first search optimization algorithm that can also be successfully used in multiplayer scenarios \cite{Kowalczyk2022DevelopingASuccessful}. 
The agent simulates only their turn, so by using large beam width it can cover most of the tree up to a given depth.
To avoid getting stuck in the local maximum, the agent uses simulated annealing to select, with a small probability, paths with lower heuristic scores.

\subsection{Comparison of Sample Agents}\label{sec:comparison}

\begin{figure}[t]
    \centering
    \includegraphics[width=\columnwidth]{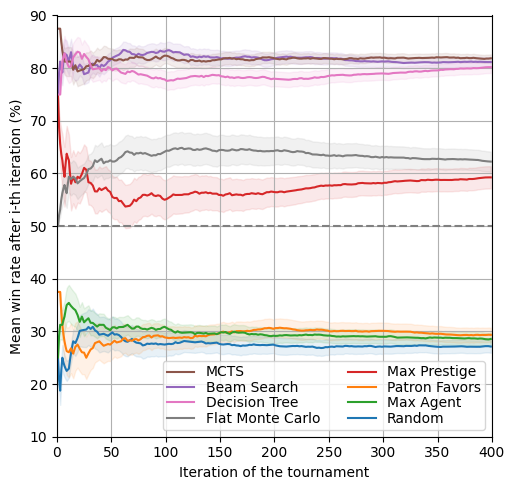}
    \caption{Results of the tournament between all sample agents.}
    \label{fig:tournament}
\end{figure}

Figure~\ref{fig:tournament} shows the results of our tournament. In each iteration, every algorithm played one game against every other. The turn limit was set to 30 seconds. We run 400 iterations in total, after which time the results seem to be relatively stable, forming three clusters of agents. 

The best three agents are clearly \textit{MCTS}, \textit{Decision Tree}, and \textit{Beam Search}. However, when we look at the scores between each pair (Table~\ref{tab:tournament}) it seems that there is no clear winner. Instead, we observe a rock-paper-scissor behavior.
This observation also indicates that a variety of techniques seem promising, and may be used to tackle the game successfully.

Although the idea behind the \textit{Max Prestige} agent is straightforward, it performs quite well, achieving significantly better scores than other simple agents. 
Also, before round 200, all agents were choosing patrons randomly, while after they switched to the Apriori algorithm, based on the win rates learned from the previous games. The only agent that visibly profited from this change is \textit{Max Prestige}.

\begin{table}[t]
    \begin{center}
        \caption{Detailed win rates (including 95\% confidence intervals) between top five agents from the tournament. 400 games between each pair.}
        \label{tab:tournament}
        \resizebox{\linewidth}{!}{%
            \begin{tabular}{c||c|c|c|c|c}
                \toprule
                versus        & Max Prestige     & Flat MC          & Beam Search               & Decision Tree             & MCTS                      \\
                \midrule
                Max Prestige  &                  & $62.7\!\pm\!4.7$ & $76.8\!\pm\!4.1$          & $\textbf{87.2}\!\pm\!3.3$ & $73.2\!\pm\!4.3$          \\
                Flat MC       & $37.2\!\pm\!4.7$ &                  & $80.8\!\pm\!3.9$          & $68.2\!\pm\!4.6$          & $\textbf{84.8}\!\pm\!3.5$ \\ 
                Beam Search   & $23.2\!\pm\!4.1$ & $19.2\!\pm\!3.9$ &                           & $\textbf{56.5}\!\pm\!4.9$ & $48.5\!\pm\!4.9$          \\ 
                Decision Tree & $12.8\!\pm\!3.3$ & $31.8\!\pm\!4.6$ & $43.5\!\pm\!4.9$          &                           & $\textbf{56.2}\!\pm\!4.9$ \\ 
                MCTS          & $26.8\!\pm\!4.3$ & $15.2\!\pm\!3.5$ & $\textbf{51.5}\!\pm\!4.9$ & $43.8\!\pm\!4.9$ &                                    \\
                \midrule
                Average       & $25.0\!\pm\!2.1$ & $21.6\!\pm\!3.5$ & $63.1\!\pm\!2.4$          & $63.9\!\pm\!2.4$ & $65.7\!\pm\!2.3$                   \\
                \bottomrule
            \end{tabular}%
        }
    \end{center}
\end{table}


\subsubsection{\ARXIV{Player bias}}

\begin{figure}[t]
    \centering
    \includegraphics[width=\columnwidth]{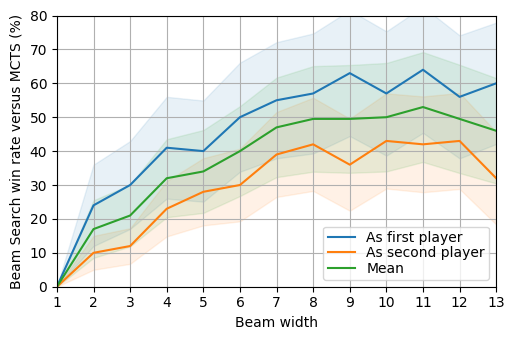}
    \caption{\ARXIV{Beam Search versus MCTS depending on the beam width and player order. As we can see, winning as the second player is significantly harder.}}
    \label{fig:beamwidth}
\end{figure}

\ARXIV{Figure~\ref{fig:beamwidth} shows the win rate of \textit{Beam Search} against \textit{MCTS} depending on the selected beam width (200 games per player order per beam width).
The difference in Beam Search win rate, depending on whether it is the first or the second player, deserves special attention, as this is not a unique behavior.}

\ARXIV{Repeating tests for other agents show that the win rate when being first is significantly higher.
Given this, during our competition, all games between agents will be played twice using the same seed, with and without switching starting positions. }

\subsection{TOTAIC 2023 Winner}

The competition organized at the IEEE Conference on Games 2023  was compatible with Tales of Tribute from ESO PC/Mac Patch 8.3.5 and contained six out of nine patrons available at that time in ESO. The \textit{Orgnum} deck was not allowed to be used by the participants. 
The prize for the winner was sponsored by IEEE Computational Intelligence Society.

The winning agent for this contest was developed by Adam Ciężkowski and Artur Krzyżyński.
The agent was based on an idea similar to root parallelization MCTS. It maintained 5 MCTS trees, each created for a different game seed, and divided the available search time between them. The final move was selected based on the average score across all trees.
Only the current turn was searched, with the heuristic evaluation function applied after the \verb|END_TURN| application. The evaluation took into account various properties of the game state and differentiated between three game stages (early, mid, late). Its feature weights were optimized using a simple evolutionary algorithm.
The scores for individual cards were calculated based on tier lists, developed by human players.
\CHANGE{The winrates of this agent against the sample agents are 79.22\% vs.\ Beam Search, 81.33\% vs.\ MCTS, and 93.12\% vs.\ Decision Tree.}
Details of the agent are described in \cite{Ciezkowski2023TotAgent}.

The source codes of submitted agents and detailed information about the results and scripts allowing for easy rerun of the tournament, are available in the repository\footnote{\url{https://github.com/ScriptsOfTribute/ScriptsOfTribute-CompetitionsArchive}}. Similar data for all future competitions will also be stored there.

\section{Conclusion}\label{sec:conclusion}

This paper introduces the first Tales of Tribute AI Competition, a new card game-based AI challenge aiming to fill the gap after no longer organized Hearthstone AI Competition and Strategy Card Game AI Competition.
Our contest is based on an existing (mini)game Tales of Tribute, an activity within the MMORPG The Elder Scrolls Online featuring an original collectible deckbuilding card game.
ToT is smaller and simpler than Hearthstone, but significantly more challenging than Legends of Code and Magic used in SCGAI.
The size of the game allows it to be fully translated into a research framework, without any handicaps or simplifications, providing equal challenge to both human and AI players. 

Just like other CCG-like games, ToT characterizes by large branching factor, randomness, and hidden information; a combination of features that algorithms still struggle with.

There are never enough challenges tackling these problems from different sides, encouraging the researchers to overcome them.
Moreover, as TOTAIC is the first card-based competition based on a deck-building game, it introduces some novel point of view.
In particular, there is more long-term planning involved, as acquired cards usually influence the board multiple turns away, and player interactions are less straightforward, as we usually do not deal with the opponent cards directly.

The domain is suited to the classic adversarial search approaches, including MiniMax and Monte Carlo Tree Search, as well as optimization algorithms (Rolling Horizon Evolutionary Algorithm, Beam Search) and Neural Networks. We expect the successful solutions to be a mix of the mentioned techniques with some rule-based decisions to handle specific cases.

The current edition of TOTAIC is hosted at the IEEE Conference on Games 2024.
We are planning to further develop our framework and especially keep improving GUI, making it more user-friendly and matching the original game experience.
The secondary goal is to make our project interesting for the ESO players base who enjoy ToT, broaden the knowledge about AI to non-academia people, and give some perspective on human vs AI tournaments.
Our long-term plan is to keep up to date with the ESO version of the game, by implementing new decks, and providing balance changes. Thus, we hope that the difficulty of the challenge will slightly advance each year, matching the increased experience of the competitors.

More information on how to participate in the Tales of Tribute AI Competition, our framework, and bot-making tutorials can be found at:  {{\texttt{\href{https://github.com/ScriptsOfTribute}{github.com/ScriptsOfTribute}}}}.

\bibliographystyle{IEEEtran} 
\bibliography{bibliography} 

\end{document}